\pdfoutput=1

\documentclass{article}

\usepackage{microtype}
\usepackage{graphicx}
\usepackage{subfigure}
\usepackage{booktabs} 
\usepackage{color}
\usepackage{tabularx}
\usepackage[skip=1ex]{caption}

\usepackage{hyperref}

\usepackage[accepted]{icml2018}

\icmltitlerunning{Information Planning for Text Data}

\begin{document}

\twocolumn[
\icmltitle{Information Planning for Text Data}



\icmlsetsymbol{equal}{*}

\begin{icmlauthorlist}
\icmlauthor{Vadim Smolyakov}{mit}
\end{icmlauthorlist}

\icmlaffiliation{mit}{Massachusetts Institute of Technology, Cambridge, USA}
\icmlcorrespondingauthor{Vadim Smolyakov}{vss@csail.mit.edu}

\icmlkeywords{Machine Learning, ICML}

\vskip 0.3in
]



\printAffiliationsAndNotice{\icmlEqualContribution} 

\begin{abstract}
Information planning enables faster learning with fewer training examples. It is particularly applicable when training examples are costly to obtain. This work examines the advantages of information planning for text data by focusing on three supervised models: Naive Bayes, supervised LDA and deep neural networks. We show that planning based on entropy and mutual information outperforms random selection baseline and therefore accelerates learning. 
\end{abstract}

\section{Introduction}
\label{intro}

Information planning involves making decisions based on information measures \cite{CoverBook}. Information planning is closely related to active learning \cite{Settles2009}, \cite{Olsson2009} and optimum experiment design \cite{FederovBook}, \cite{Chaloner1995} in which labeled data is expensive to obtain. The key idea behind information planning is that a model can learn better with fewer training examples if the training examples are carefully selected to maximize the information gain for a particular task. In other words, given an abundance of unlabelled data, we would like to rank the unlabelled examples according to their usefulness in training of the model. The top $K$ most informative training examples are labelled by expert annotators and added to the training set.

There are three main scenarios in which the learner may be able to ask queries: membership queries, stream-based selective sampling and pool-based sampling. In membership query, the learner may request labels for any unlabelled instance in the input space. In stream-based selective sampling, the active learning is done sequentially as each unlabelled instance is drawn one at a time from the data source and the learner must decide whether to query or discard it. Finally, in pool-based sampling, the entire set of unlabelled data is ranked and the most informative subset is labelled and added to the training set. In this work, we focus on pool based sampling as summarized in Algorithm \ref{alg:al}.

\begin{algorithm}[t]
\caption{Generic Information Planning}
\label{alg:al}
\begin{algorithmic}[1]
\STATE \textbf{Input:} data $x \in \{D_{train}, D_{pool}\}$, model $M$, acquisition function $a(x, M)$ 
\STATE \textbf{Output:} trained model $M^{(T)}$ 
\STATE \textbf{for} $t = 1,2,...,T$ \textbf{do}
\STATE ~~~ train $M^{(t)}$ on $D_{train}$ 
\STATE ~~~ select $x^{\ast} = \arg \max_{x \in D_{pool}} a(x, M^{(t)})$ 
\STATE ~~~ update $D_{train} = D_{train}\cup x^{\ast}$; $D_{pool} = D_{pool} \setminus x^{\ast}$
\STATE \textbf{end for}
\end{algorithmic}
\end{algorithm}

The decisions whether or not to query instances of unlabelled examples are based on a measure of informativeness or a query strategy. The simplest and most commonly used query framework is \textit{uncertainty sampling} \cite{Lewis1994}. In this framework, an active learner queries instances about which it is least certain about the label. For example, in a binary classification task, uncertainty sampling queries points near the decision boundary where the class probabilities are close to $1/2$, i.e. highest entropy. Uncertainty sampling also works for regression problems, in which case the learner queries points with highest output variance in their prediction. In general, uncertainty sampling chooses a subset $x^{\ast}$ from a set of unlabelled examples $x \in D_{pool}$ that maximizes label entropy:
\begin{equation}
    x^{\ast} = \arg\max_{x\in D_{pool}} -\sum_c p(y=c|x)\log p(y=c|x)
\end{equation}
An alternative measure that not only looks at the uncertainty of the label but also at the informativeness is mutual information:
\begin{equation}
    I(X;Y) = H(X) - H(X|Y) = H(Y) - H(Y|X)
\end{equation}
For random variables $X$ and $Y$, to maximize the \textit{information gain} $I(X;Y)$, we want to maximize $H(X)$ (uncertainty about $X$) and minimize $H(X|Y)$, i.e. we want $Y$ to be informative about $X$. The choice of random variables $X$ and $Y$ depends on the specific application and probabilistic model. As we will see, in complex models, often one direction of mutual information is easier to compute compared to the other.  

Other query selection strategies include ensemble methods such as query-by-committee \cite{Seung1992} in which an ensemble of diverse and accurate models are trained on the labeled dataset. Each ensemble member is then allowed to vote on the labels of query candidates and the most informative query is considered to be an instance about which they most disagree. In this work, we focus on uncertainty sampling and mutual information as the primary query strategies.  

In application to natural language processing, active learning has been successfully applied to information extraction \cite{Scheffer2001}, named entity recognition \cite{Becker2005}, text categorization \cite{Tong2002}, part-of-speech tagging \cite{Ringger2007} and parsing \cite{Becker2005b}. In this work, we study the benefits of information planning for text data on three different tasks: text classification, supervised topic modeling and sentiment classification. \footnote{all code is available on-line} To demonstrate the advantage of information planning, we compare performance of active learning against random selection. As the size of unlabelled dataset increases, we expect active learning to perform better.

The rest of the paper is organized as follows. Section \ref{naive_bayes} discusses information planning in Naive Bayes model for text classification. Section \ref{supervised_lda} focuses on active learning with supervised LDA. Section \ref{deep_neural_network} applies information planning to deep neural networks for sentiment prediction. Section \ref{discussion} discusses the results and the paper is concluded in Section \ref{conclusion}.


\section{Naive Bayes}
\label{naive_bayes}

\begin{figure}[thpb]
    \centering
    \includegraphics[width=0.45\textwidth]{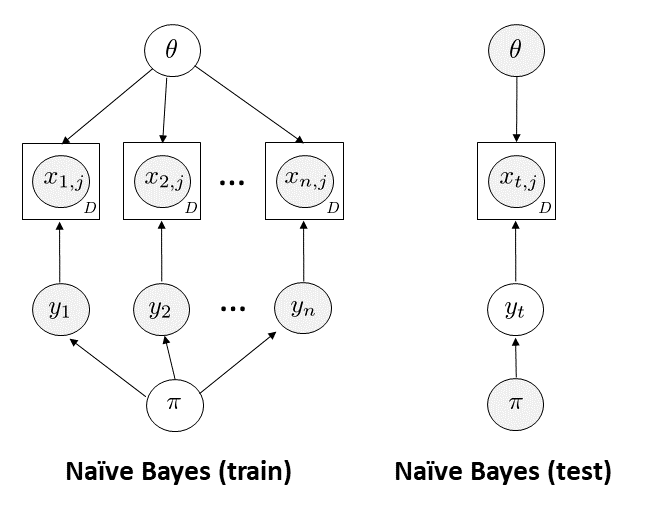}
    \caption{Naive Bayes graphical model.}
    \label{fig:nb_gm}
\end{figure}

We consider the task of text classification using a Naive Bayes graphical model as shown in Figure \ref{fig:nb_gm}. 

Let $x_{ij}$ be Bernoulli random variables indicating the presence ($x_{ij}=1$) or absence ($x_{ij}=0$) of a word $j \in \{1,...,D\}$ in document $i \in \{1,...,n\}$, parameterized by $\theta_{jc}$ for a given class label $y_i \in \{1,...,C\}$. In addition, let $\pi$ be a Dirichlet distribution representing the prior over the class labels. Thus, the total number of parameters is $|\theta| + |\pi| = \mathcal{O}(DC) + \mathcal{O}(C) = \mathcal{O}(DC)$. Due to the small number of paramters, the Naive Bayes model is immune to over-fitting \cite{MurphyML}.

The choice of Bernoulli Naive Bayes formulation is important because it leads itself to word-based information planning. By associating each word in the dictionary with a binary random variable, we are able to compute the influence of individual words on class label distribution.

Consider words $x_i$ in a single document $i$:
\begin{eqnarray}
    p(x_i,y_i|\theta) &=& p(y_i|\pi)\prod_{j=1}^{D}p(x_{ij}|y_i,\theta) \\  
    &=& \prod_{c=1}^{C}\pi_{c}^{1[y_i=c]}\prod_{j=1}^{D}\prod_{c=1}^{C}p(x_{ij}|\theta_{jc})^{1[y_i=c]} \nonumber
\end{eqnarray}

We can write-down the Naive Bayes inference algorithm by maximizing the log-likelihood objective:
\begin{eqnarray}
   &&\log p(D|\theta) = \log \prod_{i=1}^{n}p(x_i,y_i|\theta) = \sum_{i=1}^{n}\log p(x_i, y_i|\theta) \nonumber \\
   &=& \sum_{c=1}^{C}N_c \log \pi_c + \sum_{j=1}^{D}\sum_{c=1}^{C}\sum_{i:y_i=c} \log p(x_{ij}|\theta_{jc}) 
\end{eqnarray}

During test time, we would like to predict the class label $y$ given the training data $D$ and the learned model parameters. Applying the Bayes rule:
\begin{eqnarray}
    &&p(y=c|x_{i,1},...,x_{i,D}, D) \propto \nonumber \\
    &&p(y=c|D) \prod_{j=1}^{D}p(x_{ij}|y=c,D) 
\end{eqnarray}
Substituting $p(x_{ij}|y=c, D)$ and taking the log, we get:
\begin{eqnarray}
    &&\log p(y=c|x,D)\propto \log p(y=c|D) + \\
    &&\sum_{j=1}^{D}\big(1[x_{ij}=1]\log \theta_{jc} + 1[x_{ij}=0]\log(1-\theta_{jc}) \big) \nonumber
\end{eqnarray}

\subsection{Information Planning}

In entropy based planning we want to rank test documents according to the entropy of their class label. In the case of class distribution, the entropy can be computed in closed form: $H(y) = -\sum_{c=1}^{C}p(y=c) \log p(y=c)$.

In the case of mutual information, we need to choose the random variables we want to use for planning. Since we are interested in classifying documents, it makes sense to choose $y_i$ (the class label for document $i$) as one of the variables. Our choice of the second variable depends on the quantity of interest. We can choose one of the global variables $\theta_{jc}$ (probability that word $j$ is present in class $c$) or $\pi_c$ (probability of class $c$). In this example, we consider estimating $I(y_i;\theta)$, i.e. we are interested in measuring the information gain about the word distribution $\theta$ given the class label $y_i$ for a test document. Since both variables are discrete, the mutual information can be estimated in closed form:
\begin{eqnarray}
    &&I(y_i;\theta) = \mathrm{KL}\big(p(y_i,\theta)||p(y_i)p(\theta)\big) = \nonumber \\
    &&\sum_{y \in C}\sum_{\theta} p(\theta|y_i)p(y_i)\log \frac{p(\theta|y_i)}{p(\theta)}
\end{eqnarray}
We can compute $p(\theta) = \sum_{c\in C} p(x_j=1|y=c)p(y=c)$ for $j\in\{1,...,D\}$. In addition, we compute $I(x_j;\pi)$ to measure how informative each word $x_j$ is to the global label distribution $\pi$. 

\begin{figure}[b]
    \centering
    \includegraphics[width=0.45\textwidth]{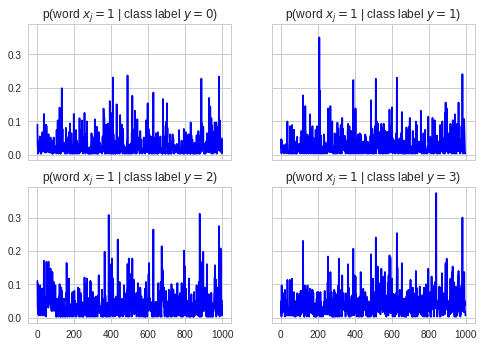}
    \caption{Naive Bayes class-conditional posterior.}
    \label{fig:nb_topics}
\end{figure}

We train our Naive Bayes classifier on a subset of the 20newsgroups dataset. In particular, we'll restrict ourselves to 4 classes: space, graphics, autos and hockey. We'll use a count vectorizer to produce a vector of word counts for each document while filtering stop and low frequency words. Figure \ref{fig:nb_topics} shows the posterior word distribution for the four classes.

\begin{figure}[b]
    \centering
    \includegraphics[width=0.45\textwidth]{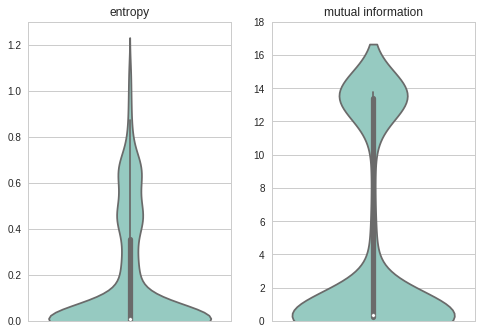}
    \caption{Violin plot showing the density of entropy and MI on test data.}
    \label{fig:nb_info1}
\end{figure}

\begin{figure}[h]
    \centering
    \includegraphics[width=0.45\textwidth]{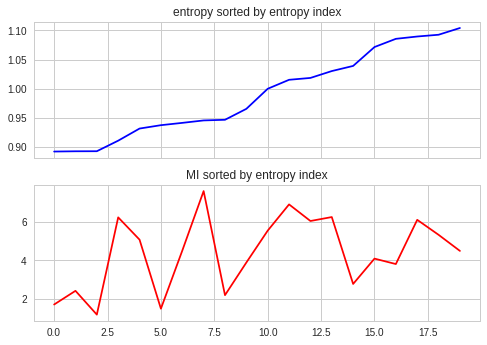}
    \caption{Test document entropy and MI sorted by entropy index.}
    \label{fig:nb_info2}
\end{figure}

Let's visualize the test document ranking based on entropy and mutual information in Figure \ref{fig:nb_info1}.
In particular, we are interested in outliers, i.e. documents that have highest entropy and mutual information. To visualize the difference between entropy and MI based planning, we can sort the documents according to entropy and use the same index to sort MI as shown in Figure \ref{fig:nb_info2}.
In Figure \ref{fig:nb_info2} documents that have high entropy but low MI are uncertain but not informative. Therefore, we expect MI to provide a better measure for information planning.

\begin{figure}[h]
    \centering
    \includegraphics[width=0.45\textwidth]{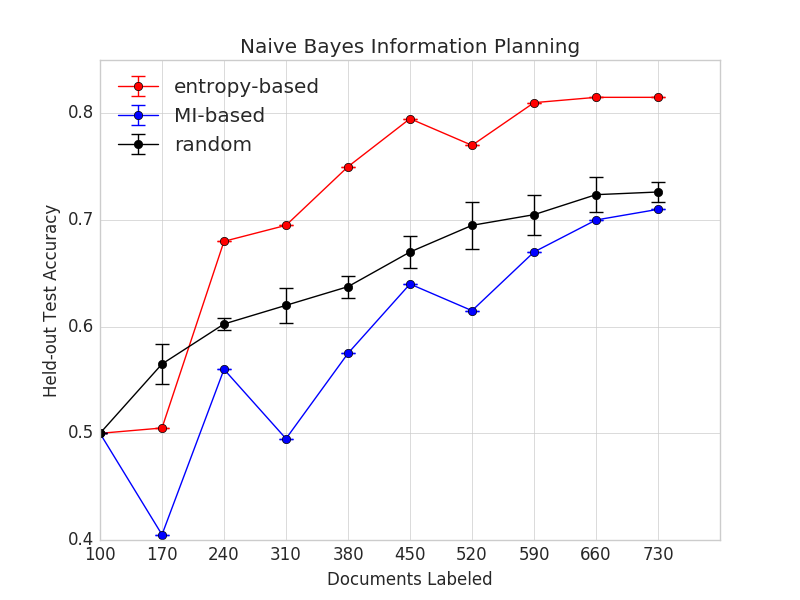}
    \caption{Classification accuracy as a function of labeled examples.}
    \label{fig:nb_trials_acc}
\end{figure}

To evaluate the utility of information planning in comparison to random selection, we take a total of $1K$ documents and divide them into $100$ labelled training documents, $700$ unlabelled test documents, and holdout $200$ documents for evaluating classification performance. We run the active learning pipeline, in which top $K$ most informative documents are selected from the test set, labelled by annotators and added to the training set, at which point the modeled is re-trained from scratch and evaluated on the $200$ heldout documents. Figure \ref{fig:nb_trials_acc} shows the experimental results of this active learning pipeline.
We can see a clear advantage of using information planning compared to random selection.

\section{Supervised LDA}
\label{supervised_lda}

\begin{figure}[thpb]
    \centering
    \includegraphics[width=0.45\textwidth]{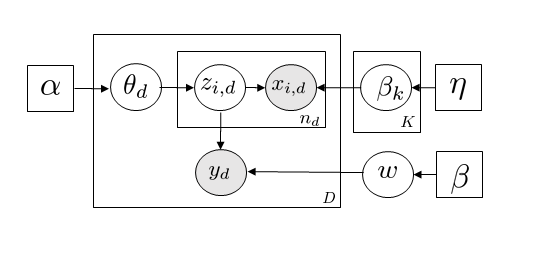}
    \caption{Supervised LDA \cite{Blei2010} graphical model.}
    \label{fig:slda_gm}
\end{figure}

Latent Dirichlet Allocation (LDA) \cite{Blei2003} is an unsupervised topic model that represents each document as a mixture of topics, where a topic is a distribution over words. The objective is to learn the shared topic distributions and their proportions for each document. However, it is often desirable to associate a quantity of interest (e.g. a score) with each document; this could be a five star rating for a movie review, a quality score of a report, or a number of times an on-line article is linked. Supervised LDA (sLDA) \cite{Blei2010} is designed to jointly learn the topics and the score variable associated with each document. The graphical model of supervised LDA is shown in Figure \ref{fig:slda_gm}. 

\begin{figure}[h]
    \centering
    \includegraphics[width=0.45\textwidth]{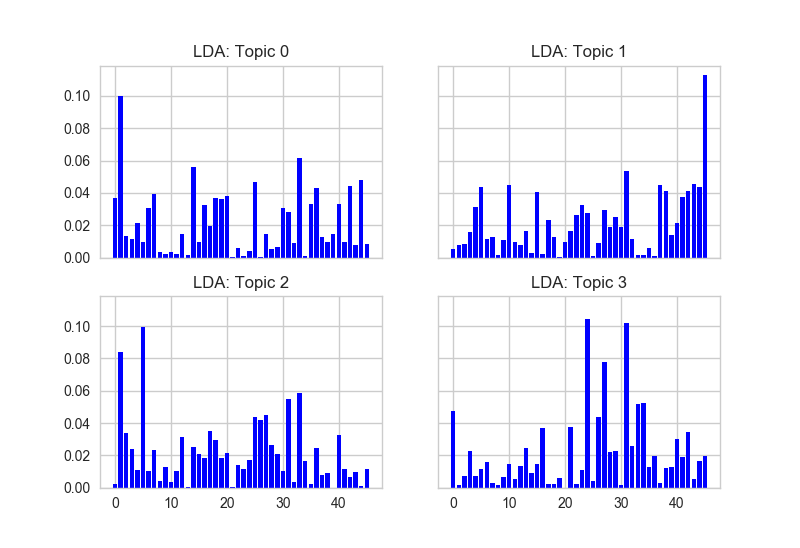}
    \caption{Supervised LDA topics inferred using PyMC \cite{PyMC2010}.}
    \label{fig:slda_topics}
\end{figure}

sLDA associates each word $x_{id}$ with a topic label $z_{id}\in \{1,...,K\}$. Each document is associated with topic proportions $\theta_d$ that can also be used as a measure of document similarity. The topics $\beta_k$ represented by the Dirichlet distribution are shared across all documents. The hyper-parameters $\alpha$ and $\eta$ capture our prior knowledge of topic proportions and the topics, e.g. from past training of the model. Finally, $y_d$ is the response variable that represents a score associated with each document. In this example, $y_d$ is modelled as a regression between global coefficients $w$ and a normalized histogram of topics present in a document: $\bar{z}$. Assuming a Gaussian distribution, $y_d$ can be expressed as follows:
\begin{eqnarray}
   y_d &\sim& N(w_1\bar{z}_1 + ... + w_k\bar{z}_k, \sigma^2) \\
   \bar{z} &=& \frac{1}{N}\sum_{n=1}^{N}z_n
\end{eqnarray}
where $z_n$ is a one-hot encoded topic assignment vector of size $1\times K$. We use PyMC \cite{PyMC2010} Metropolis-Hastings sampler to infer the latent variables. Figure \ref{fig:slda_topics} shows the first $4$ (out of $8$) topics of sLDA model trained on the polarity sentiment dataset \cite{PolarityDataset}. Notice, that the addition of the score variable $y_d$ influences topic inference due to the coupling between $w$ and $\beta_k$ introduced by the $z_{i,d}\rightarrow y_d \leftarrow w$ and $z_{i,d}\rightarrow x_{i,d} \leftarrow \beta_k$ v-structures \cite{KollerBook}. 


\subsection{Information Planning}

We can use a sample based estimator to compute the entropy of the score variable $y_d$: $H[y_d|x_d, D] = E[-\log p(y_d|x_d, D)]$.

Assuming we have topic samples:
\begin{eqnarray}
   && \{(z_{1d}^{(i)}, ..., z_{Nd}^{(i)})\}_{i=1}^{M} \stackrel{MCMC}{\sim} p(z_{1d},...,z_{Nd}|x_d, D) \\
   && p(y_d|x_d, D) = \sum_{z_{1d}}...\sum_{z_{Nd}} p(z_{1d},...,z_{Nd}|x_d, D)p(y_d|z_d)\nonumber \\
   && \approx \frac{1}{M} \sum_{i=1}^{M}p(y_d|z_d = z_{d}^{(i)}) = \hat{p}(y_d|x_d, D)
\end{eqnarray}
Once we have a sample estimate $\hat{p}(y_d|x_d, D)$, to compute entropy, we need to estimate the expectation:
\begin{eqnarray}
   &&E[-\log \hat{p}(y_d|x_d,D)] = \nonumber \\
   &&= - \int p(y_d|x_d, D) \log \hat{p}(y_d|x_d, D) dy_d \nonumber \\
   &&\approx \frac{1}{M}\sum_{i=1}^{M}\log \hat{p}(y_{d}^{(i)}|x_d, D)
\end{eqnarray}
where $\{(y_{d}^{(i)})\}_{i=1}^{M} \stackrel{MCMC}{\sim} \hat{p}(y_{d}^{(i)}|x_d, D)$.

\begin{figure}[thpb]
    \centering
    \includegraphics[width=0.4\textwidth]{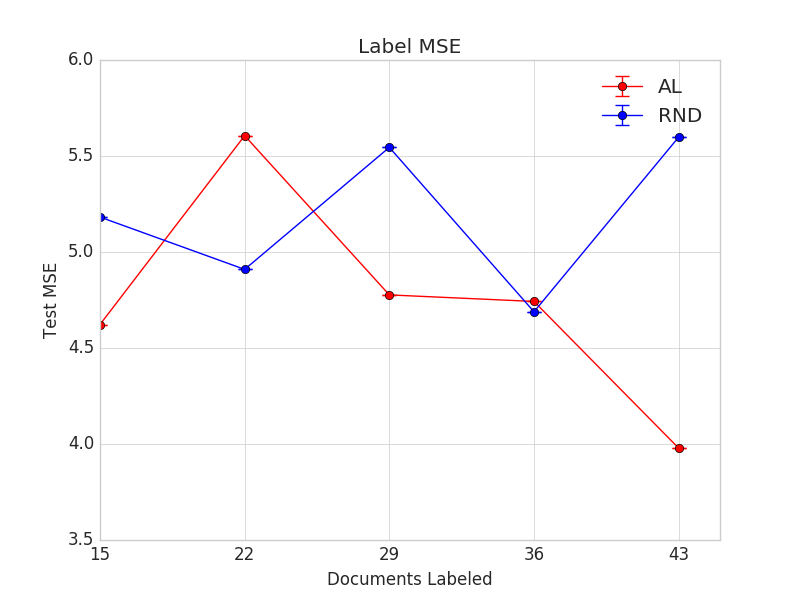}
    \caption{Supervised LDA test MSE.}
    \label{fig:slda_test_mse}
\end{figure}

To evaluate the utility of information planning in comparison to random selection, we start with a corpus of $100$ documents, from which we allocate $15$ documents for training, $35$ for testing and holdout $50$ documents for evaluating sLDA performance. Figure \ref{fig:slda_test_mse} shows the test MSE of the Gaussian score variable evaluated against the ground truth review score on the heldout set of $50$ documents.

We can see that active learning selection achieves smaller MSE in comparison to random selection baseline.


\section{Deep Neural Network}
\label{deep_neural_network}

We consider information planning in the case of deep neural networks. In particular, we examine two architectures: LSTM and CNN in application to sentence sentiment classification \cite{DLBook}. 

Recurrent Neural Networks (RNNs) provide a natural way of encoding a sentence as a sequence of words. Just as you are processing this sentence word by word while keeping a memory of what came before, RNNs maintain an internal model of the past and update it as soon as the new information arrives. Thus, RNNs contain an internal loop unrolled over the length of the input sequence. As a result, RNNs can capture more information in comparison to $n$-gram models. For example, in a $n$-gram language model (with $n=3$), we can predict the next word given the previous two:
\begin{equation}
    l_{sent} = \frac{1}{n}\log P(w_1,...,w_n) = \frac{1}{n}\sum_{i=1}^{n}\log p(w_i|w_{i-1}, w_{i-2})
\end{equation}
The dependence in neural language model (LM) extends to the beginning of the sentence, i.e. $p(w_{i}|w_{i-1},...,w_1)$, while in a $n$-gram LM the conditioning is only on the previous $n$ words and as $n$ grows larger, sparsity of training data becomes a problem. In both cases, the word conditional distribution greatly reduces the output space of possible words and enables learning of word representations based on their context \cite{word2vec2013}.    

\begin{table*}[t]
\centering
\begin{tabularx}{\linewidth}{XX}
    \includegraphics[width=\columnwidth]{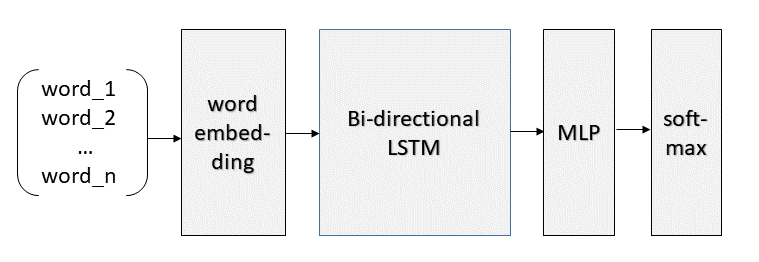}
    \captionof{figure}{LSTM model architecture}
    \label{fig:lstm_arch}
&
    \includegraphics[width=\columnwidth]{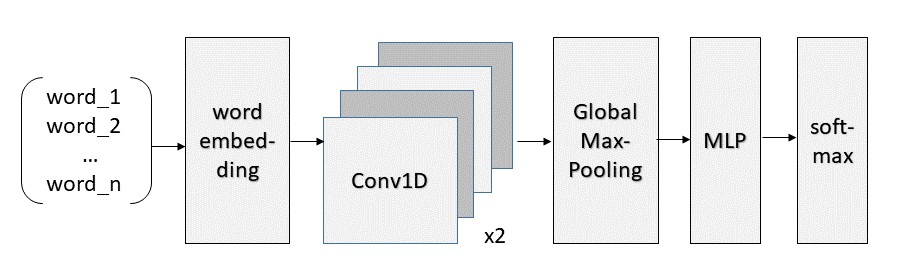}
    \captionof{figure}{CNN model architecture}
    \label{fig:cnn_arch}
\\
\end{tabularx}
\end{table*}

A sentence can be encoded by an RNN (such as LSTM) as either the output of its last layer or via average pooling of its intermediate outputs. In both cases, the word position information is preserved in contrast to averaging of word vectors. Figure \ref{fig:lstm_arch} shows the LSTM architecture used for sentiment classification. We use a bi-directional LSTM that concatenates sequence representations of forward and reverse LSTMs. This enriches order-sensitive representation of the sentence and leads to a slight increase in accuracy. In addition to output layer dropout, we use recurrent dropout that applies the same dropout mask at every time-step (in contrast to a dropout mask that varies randomly between time-steps that could harm the learning process) \cite{Gal2016}.   

Convolutional Neural Networks (CNNs) offer computational advantages due to their parallel nature and comparable performance with time-series data, especially in cases when recent past is not more informative than the distant past (due to translation invariance of CNNs). The same properties that make convnets excel at computer vision also make them highly relevant for sequence processing. Time can be treated as a spatial dimension like the height or width of a 2D image. For example, 1D convolutions can be used to extract patterns from local 1D patch sub-sequences of the input sequence. Because the same transformation is performed on every patch, a pattern learned at a certain position in the sentence can later be recognized at a different position, making 1D convnets translation invariant. While a word-level conv-net operating on word embeddings with a kernel size of $5$ should be able to recognize words or word fragments of length $5$ or less, a character-level 1D conv-net is able to learn about word morphology. In addition, we can learn hierarchical patterns by stacking several convolutional layers. Figure \ref{fig:cnn_arch} shows the CNN architecture used for sentiment classification. It consists of a word embedding layer followed by a stack of two convolutional layers separated by a dropout layer. Finally, the activations across each of the filters are pooled via global max pooling and fed into a Multi-Layer Perceptron (MLP) classifier.  

For both LSTM and CNN models, we choose pre-trained word embeddings because in an active learning setting, there's usually not enough training data to learn specialized word embeddings for a given task. In order to preserve the latent structure (that could otherwise be destroyed by large gradients from randomly initialized downstream layers), we configure the embeddings to be non-trainable. Both architectures use Glove pre-trained word embeddings \cite{Glove2014} based on word co-occurence for dense representation of words. 


\subsection{Information Planning}

As shown in \cite{MCDropout2016} dropout \cite{Dropout2014} and other stochastic regularization techniques can be used for approximate inference in deep neural networks. This is accomplished by using dropout at test-time and computing Monte Carlo integration (referred to as MC dropout):
\begin{eqnarray}
    &&p(y=c|x, D) = \int p(y=c|x, w)p(w|D)dw \\ 
    &&\approx \int p(y=c|x,w)q_{\theta}^{\ast}(w)dw \approx \frac{1}{T}\sum_{t=1}^{T}p(y=c|x,\hat{w}_t) \nonumber
\end{eqnarray}
where $\hat{w}_{t} \sim q_{\theta}^{\ast}(w)$ and $q_{\theta}^{\ast}(w)$ is a variational distribution that minimizes KL divergence to the true model posterior $p(w|D)$:
\begin{equation}
    \arg \min_{\theta} KL(q_{\theta}(w) || p(w|D))
\end{equation}
where $\theta$ are the variational parameters and $w$ are the neural network weights. Averaging forward passes with MC dropout during test time is equivalent to Monte Carlo integration over a Gaussian process posterior approximation \cite{Gal2017}. 

We consider two acquisition functions: one based on entropy and one based on mutual information. An \textit{acquisition function} $a(x,M)$ determines which measurements to take next, and it's a function of the pool of unlabelled examples $x \in D_{pool}$ and the model $M$:
\begin{equation}
    x^{\ast} = \arg \max_{x \in D_{pool}} a(x,M)
\end{equation}
In the case of entropy, we want to choose data points that maximize:
\begin{equation}
    H[y|x, D] = -\sum_{c}p(y=c|x,D)\log p(y=c|x,D)
\end{equation}
In the case of mutual information, we want to choose data points that maximize the information gain between the predictions $y$ and the model parameters $w$:
\begin{equation}
    I(y;w|x,D) = H[y|x, D] - E_{p(w|D)}[H[y|w,x,D]]
\end{equation}
where $w$ in our case are the weights of the neural network and the expectation is computed using MC dropout. Thus, we are seeking data points $x$ for which the model is both marginally most uncertain about $y$ and most confident about the predicted label. Notice, that it's easier to compute MI as written above (as opposed to its alternative version) because the entropies are computed in a discrete, low-dimensional output space. This interpretation allows reasoning about uncertainty in deep learning and the application of Bayesian methods to analyze existing deep learning frameworks. 

We evaluate the effectiveness of information planning on Kaggle movie review sentiment dataset 
\cite{KaggleSentiment}. The dataset consists of phrases from the Rotten Tomatoes dataset classified into $5$ sentiment labels: negative, somewhat negative, neutral, somewhat positive and positive. Obstacles such as sentence negation, sarcasm, and language ambiguity make this a challenging task. We take the first 10K sentences, set aside 10\% for training and holdout 10\% for evaluating the performance. At each iteration, we select the top $K=800$ most informative sentences to be annotated and added to the training set. Figures \ref{fig:cnn_acc_trials}, \ref{fig:cnn_entropy_trials}, \ref{fig:lstm_acc_trials}, \ref{fig:lstm_entropy_trials} show active learning experimental results.

\begin{table*}[t]
\centering
\begin{tabularx}{\linewidth}{XXXX}
    \includegraphics[width=0.55\columnwidth]{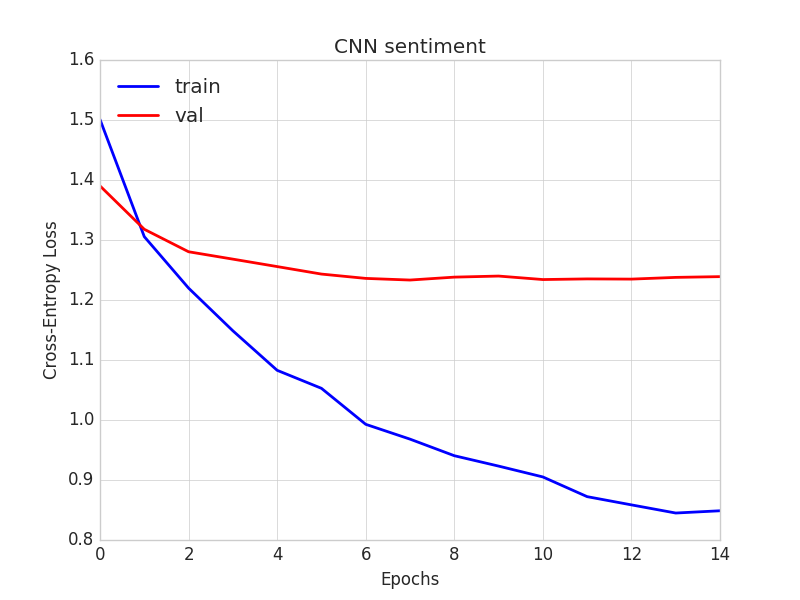}
    \captionof{figure}{CNN: training loss}
    \label{fig:cnn_loss}
&
    \includegraphics[width=0.55\columnwidth]{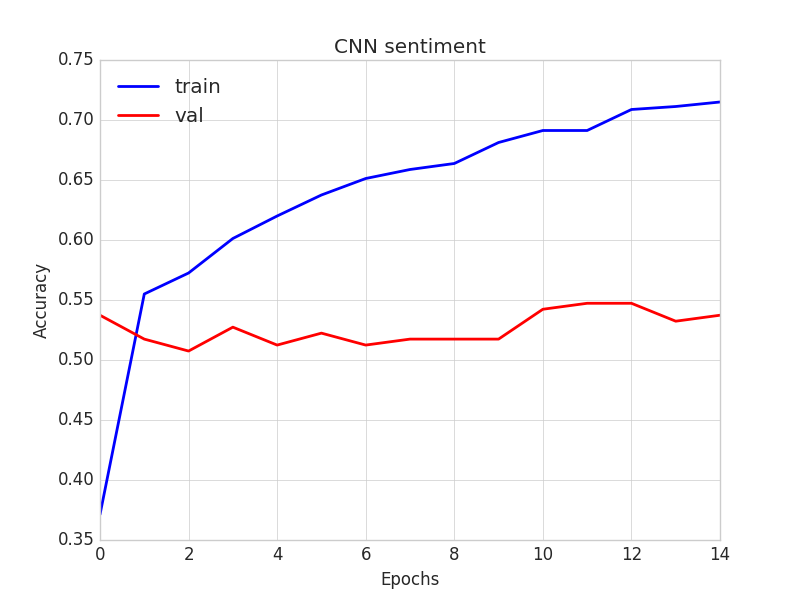}
    \captionof{figure}{CNN: training acc}
    \label{fig:cnn_acc}
&
    \includegraphics[width=0.55\columnwidth]{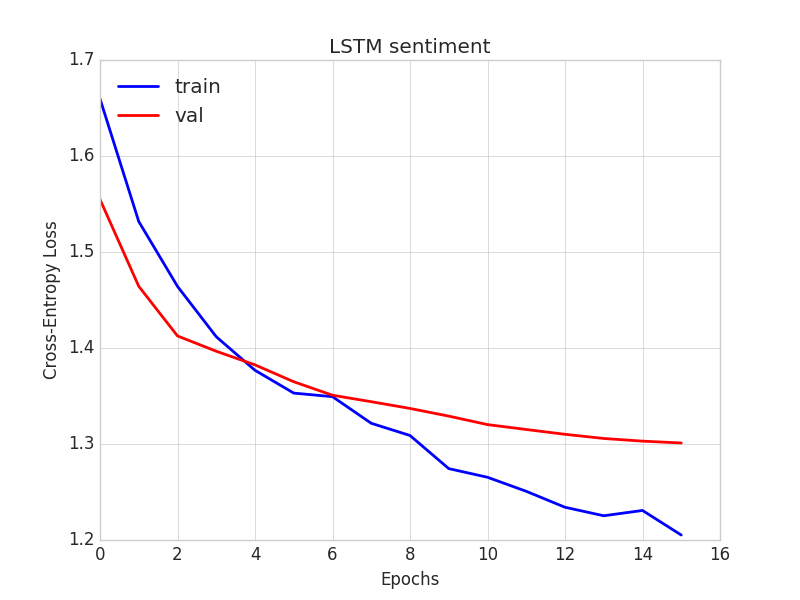}
    \captionof{figure}{LSTM: training loss}
    \label{fig:lstm_loss}
&
    \includegraphics[width=0.55\columnwidth]{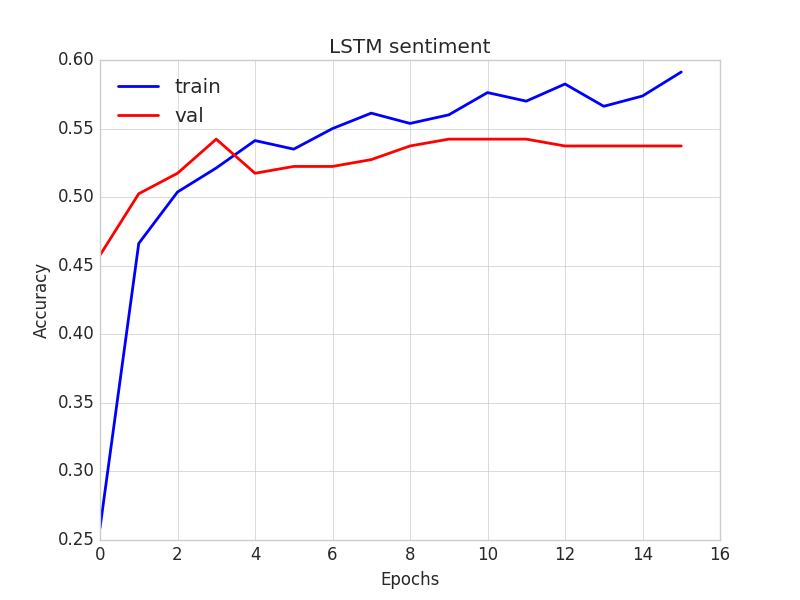}
    \captionof{figure}{LSTM: training acc}
    \label{fig:lstm_acc}
\\
\end{tabularx}
\end{table*}
   
\begin{table*}[t]
\centering
\begin{tabularx}{\linewidth}{XXXX}
    \includegraphics[width=0.55\columnwidth]{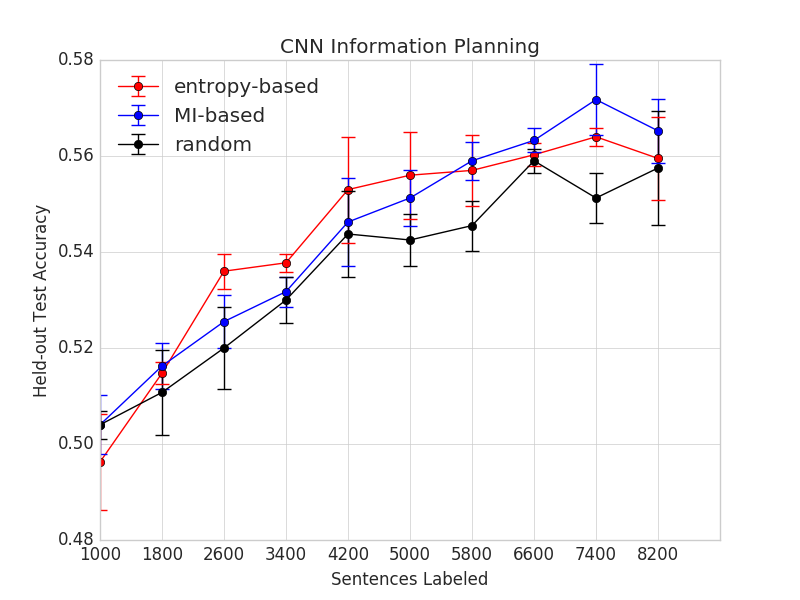}
    \captionof{figure}{CNN: test accuracy}
    \label{fig:cnn_acc_trials}
&
    \includegraphics[width=0.55\columnwidth]{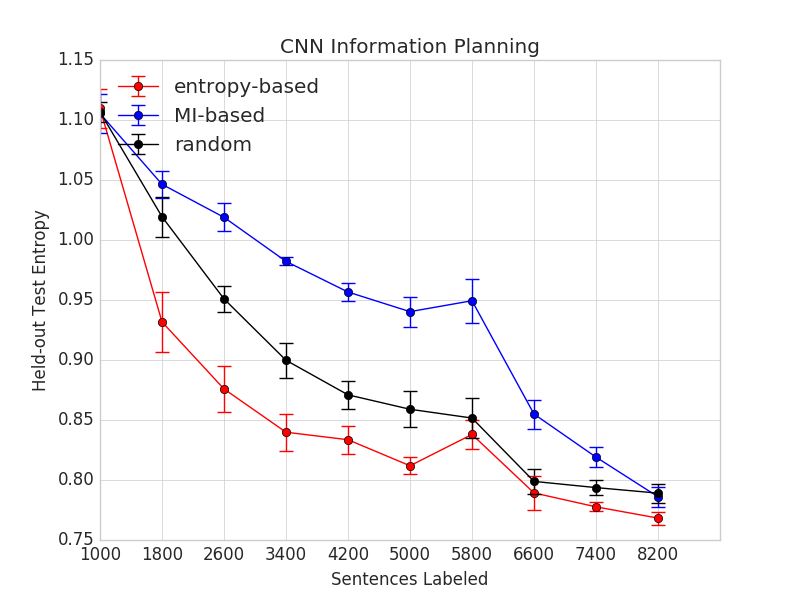}
    \captionof{figure}{CNN: test entropy}
    \label{fig:cnn_entropy_trials}
&
    \includegraphics[width=0.55\columnwidth]{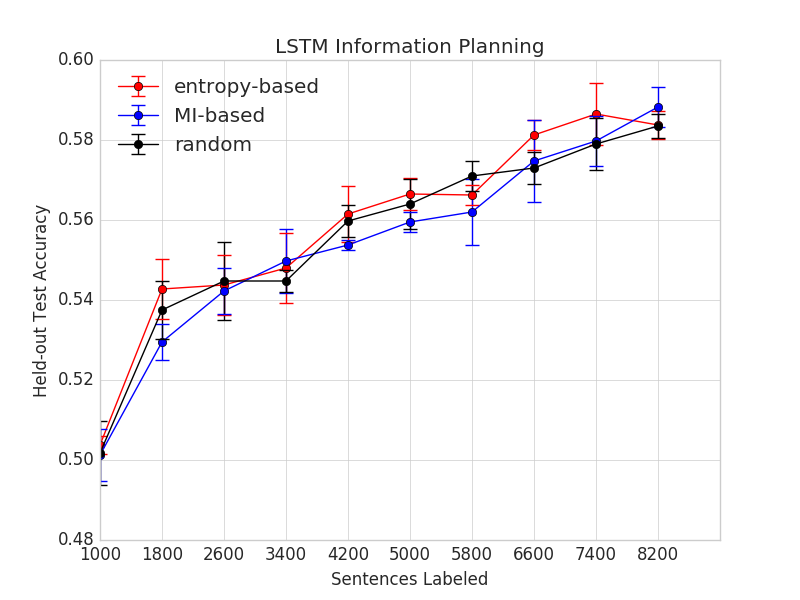}
    \captionof{figure}{LSTM: test acc.}
    \label{fig:lstm_acc_trials}
&
    \includegraphics[width=0.55\columnwidth]{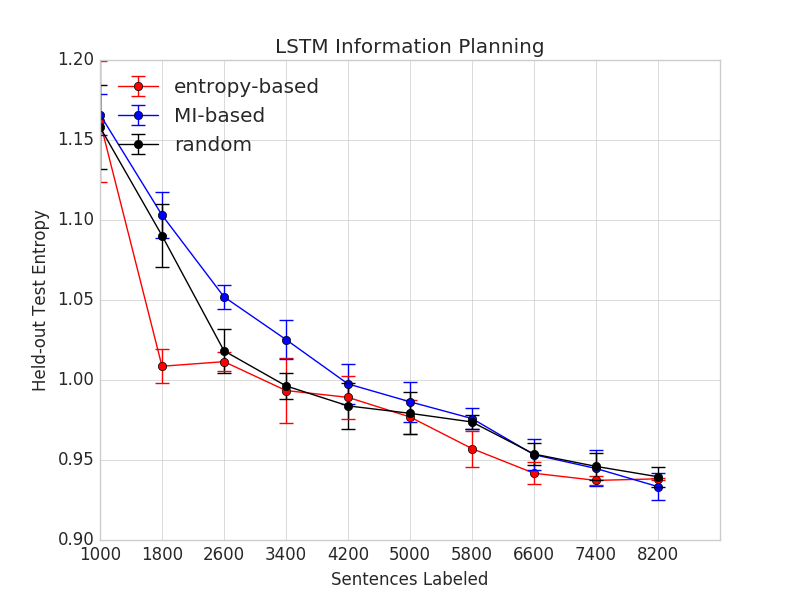}
    \captionof{figure}{LSTM: test entropy}
    \label{fig:lstm_entropy_trials}
\\
\end{tabularx}
\end{table*}
   
\section{Discussion}
\label{discussion}
In this section we discuss information planning results for each model.

\subsection{Naive Bayes}
Naive Bayes is the simplest model for studying information planning since all latent variables are discrete and information measures can be computed in closed form. By choosing the Bernoulli formulation of Naive Bayes we were able to do planning at the word level. In particular, the top-10 highest MI words as shown in Table \ref{tab:nb_words} were found to be informative of the ground truth labels.  
\begin{table}[h]
\begin{center}
  \begin{tabular}{c|c|c|c|c}
    \toprule
    class 0 & class 1 & class 2  & class 3 & MI       \\ \midrule
    get     & think   & first    & nasa    & teams    \\
    please  & well    & time     & could   & program  \\
    use     & know    & season   & much    & league   \\
    like    & also    & year     & know    & moon     \\
    one     & cars    & play     & get     & playoffs \\
    anyone  & get     & hockey   & also    & orbit    \\
    thanks  & like    & one      & like    & players  \\
    graphics& one     & would    & one     & earth    \\
    would   & would   & game     & would   & games    \\
    know    & car     & team     & space   & nasa     \\
    \bottomrule
  \end{tabular}
\end{center}
\caption{Naive Bayes: top 10 words. Ground truth labels: graphics, autos, hockey, space.}
\label{tab:nb_words}
\end{table}

From Figure \ref{fig:nb_trials_acc}, we can see that we achieve about $10\%$ accuracy increase when we use uncertainty sampling in comparison to random selection. All experimentes were repeated $10$ times.  

\subsection{Supervised LDA}

In supervised LDA model, the score variable is learned jointly with the topics. Figure \ref{fig:slda_test_mse} shows that smaller MSE is achieved with uncertainty sampling in comparison to random selection. We also note that sampling based inference for sLDA is more suitable to stream queres in which documents are queried one at a time due considerable time it takes to run sampling for the entire pool of unlabelled examples. 


\subsection{Deep Neural Networks}

In deep neural network setting we assume that computational expense of re-training the model from scratch is justified by the cost of obtaining the labels. From the training and validation loss in Figures \ref{fig:cnn_loss}, \ref{fig:lstm_loss}, we see no signs of over-fitting due regularization with dropout and weight decay. The validation loss plateaus early in the training indicating that model capacity can be increased in an attempt to achieve higher classification accuracy. Figures \ref{fig:cnn_acc}, \ref{fig:lstm_acc} show that CNN and LSTM model achieve comparable accuracy on the sentiment classification task with validation accuracy close to $55\%$.     
Figures \ref{fig:cnn_acc_trials}, \ref{fig:lstm_acc_trials} show the benefit of information planning over the random basline. We can see that we achieve close to $1\%$ improvement in accuracy in the case of CNN and a relatively smaller improvement in the case of LSTM, even thought LSTM achieves $2\%$ higher held-out test accuracy in the end. It's also interesting to see, the decrease in test label entropy as shown in Figures \ref{fig:cnn_entropy_trials}, \ref{fig:lstm_entropy_trials}, where as expected the uncertainty sampling leads to highest reduction in entropy. 
 

\section{Conclusion}
\label{conclusion}
We examined two information measures: entropy and mutual information for active learning with text data. In particular, we focused on three supervised tasks: text classification with Naive Bayes model, supervised topic modeling with sLDA, and sentiment classification with deep neural networks. In each case, we saw higher performance achieved with information planning in comparison to a random selection baseline. We hope the results of this work will encourage the use of information planning in settings where labels are expensive to obtain.  

\bibliography{paper}
\bibliographystyle{icml2018}


\end{document}